# ANOMALY DETECTION THROUGH LATENT SPACE RESTORATION USING VECTOR QUANTIZED VARIATIONAL AUTOENCODERS

*Sergio Naval Marimont [1], Giacomo Tarroni [1,2]*

[1] Cit-AI, Department of Computer Science, City, University of London
[2] BioMedIA, Department of Computing, Imperial College London

## ABSTRACT

We propose an out-of-distribution detection method that combines density and restoration-based approaches using Vector-Quantized Variational Auto-Encoders (VQ-VAEs). The VQ-VAE model learns to encode images in a categorical latent space. The prior distribution of latent codes is then modelled using an Auto-Regressive (AR) model. We found that the prior probability estimated by the AR model can be useful for unsupervised anomaly detection and enables the estimation of both sample and pixel-wise anomaly scores. The sample-wise score is defined as the negative log-likelihood of the latent variables above a threshold selecting highly unlikely codes. Additionally, out-of-distribution images are restored into in-distribution images by replacing unlikely latent codes with samples from the prior model and decoding to pixel space. The average L1 distance between generated restorations and original image is used as pixel-wise anomaly score. We tested our approach on the MOOD challenge datasets, and report higher accuracies compared to a standard reconstruction-based approach with VAEs.

*Index Terms*— Unsupervised anomaly detection, out-of-distribution, VAE, Vector Quantized-VAE,

## 1. INTRODUCTION

A wide range of methods using deep learning has been recently proposed to automatically identify anomalies in medical images [1]. Most of them are based on supervised learning, and consequently have two important constraints. First, they require large and diverse annotated datasets for training. Second, they are specific to the abnormalities annotated in the datasets, and therefore are unable to generalize to other pathologies. Unsupervised anomaly detection methods, on the other hand, aim to overcome these constraints by not relying on annotated datasets [2]. Instead, they focus on learning the underlying distribution of normal images and then identifying as anomalies the images that do not conform to the learnt distribution.

Recently, methods based on Variational Auto-Encoders (VAEs) have been proposed to identify and localize anomalies in medical images [2,3,4]. VAEs are generative models trained by minimizing a loss function composed of a reconstruction term (measuring the distance between original images and reconstructions) and a Kullback–Leibler (KL) divergence term (measuring the distance between the latent distribution and a prior, generally assumed to be Gaussian). The default approach consists in using the reconstruction loss to identify samples with anomalies, based on the assumption that the VAE will reconstruct their anomaly-free versions. However, recent results suggest that the KL divergence is actually a better anomaly score [3]. This can be caused by the high representational power of VAEs, which can reconstruct even (previously unseen) anomalies. In addition, in [4] anomalies are modelled as spatially localized deviations from a prior distribution of normal images. Gradient descent in pixel space is used to "restore" images, effectively removing anomalies. Anomalies are then localized by comparing original images to restorations. Restoration-based approaches seem to overall outperform reconstruction-based ones [2].

Under the hypothesis that abnormal images are encoded in different, lower density regions in the latent space, we propose to use an estimated latent density as anomaly score. Vector-Quantized VAEs (VQ-VAEs) [5] are well suited for this strategy because their discrete latent distribution can be modelled with expressive Auto-Regressive (AR) models, which provide state of the art performance in density estimation in images. Additionally, we enable anomaly localization relying on the generative capabilities of the AR model, with a method that we refer to as Latent Space Restoration. Results obtained in the MOOD challenge datasets (consisting of brain MR and abdominal CT images) suggest that our approach outperforms a standard reconstruction-based anomaly detection method using VAEs.

## 2. METHODS

### 2.1. Vector Quantized Variational Auto-Encoders

Vector Quantized Variational Auto-Encoders (VQ-VAEs) [5] encode observed variables in a discrete latent space instead of a continuous one. The discrete latent space can be very expressive, allowing the generation of high quality and detailed reconstructions. The discrete latent space also enables the pairing with AR models, which can independently learn the prior distribution (**Fig. 1**).

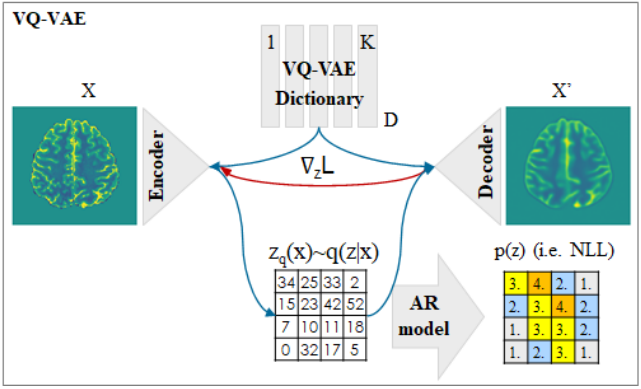

**Fig. 1.** Diagram of the proposed approach.

VQ-VAEs are built around a dictionary that maps $K$ discrete keys to a D-dimensional embedding space. In other words, the encoder network maps observed variables to the embedding space, $z_e(x) \in R^D$. Differently from standard VAEs, for which the posterior $q(z|x)$ follows a Gaussian distribution, the posterior in VQ-VAEs is categorical and deterministic, and is defined as the index of the nearest embedding vector. $e_j$, to the encoder output:

$$q(z = k|x) = \begin{cases} 1, \text{for } k = \text{argmin}_j \lVert z_e(x) - e_j \rVert_2 \\ 0, \text{otherwise} \end{cases}$$

Finally, the decoder network takes as input $z_q(x) \sim q(z|x)$ (i.e. the embedding in the dictionary nearest to the encoder output) and learns to reconstruct the observed variable distribution $p(x|z_q(x))$.

Network parameters for both encoder and decoder networks and embeddings are learnt using back-propagation. Given that the *argmin* operator is non-differentiable, the gradient in the encoder is usually approximated using *straight-through* estimator [5]. Additional terms in the VQ-VAE loss function are introduced to provide gradients to the embeddings and to incentivize the encoder to commit to embeddings. The complete VQ-VAE loss function is therefore defined as

$$L = \log\left(x|z_q(x)\right) + \lVert \text{sg}[z_e(x)] - e \rVert_2^2 + \lVert \text{sg}[e] - z_e(x) \rVert_2^2$$

where sg[.] represents the stop gradient operator.

### 2.2. Auto-Regressive prior modelling

In our method, the prior distribution of VQ-VAE is learnt using an Auto-Regressive model (AR). This will allow the estimation of the probability of samples and consequently the identification of anomalies (defined as samples associated to low probability). In addition, since AR models are generative, they enable the iterative sampling of one variable at a time, a property that we will leverage to generate multiple restorations. In an AR model, the joint probability is modelled using factorization, meaning that each variable is modelled as dependent from previous variables: $p(x) = \prod_i^N p(x_i|x_1, \dots, x_{i-1})$. In our implementation, we used the PixelSNAIL [6] architecture for the AR model.

### 2.3. Sample-wise anomaly score estimation

A sample-wise anomaly score is a numerical indicator of how likely it is for a given sample to contain an anomaly. Scores are generally either density-based (based on an estimated probability of a sample) or reconstruction-based (based on the assumption that models trained on normal data will not be able to reconstruct anomalies).

Consistent with our previous findings using VAEs, VQ-VAE with a large enough latent space are able to reconstruct abnormal regions of images and this makes the full VQ-VAE loss a poor anomaly score. However, we found that abnormal regions translate into unusual latent variables, for which the AR model assigns low probability. Therefore, we derived a sample wise anomaly score from the prior probability estimated by the AR model.

A negative log-likelihood (NLL) threshold $\lambda_s$ defines highly unlikely latent variables. The proposed sample-wise anomaly score ($AS_{\text{sample}}$) is the sum of NLL of the latent variables above threshold (over a total of $N$ variables):

$$AS_{sample} = \sum_i^N \xi(p(x_i))$$

$$\xi(z) = \begin{cases} -\log(z), \text{if } -\log(z) > \lambda_s \\ 0, \text{otherwise} \end{cases}$$

### 2.4. Pixel-wise anomaly score estimation

Pixel-wise anomaly scores quantify, for each pixel in an image, its likelihood of containing an anomaly, consequently providing anomaly localization.

The proposed pixel-wise score follows the restoration paradigm presented in [4]. Our restoration method consists in replacing high loss latent variables with samples from the learnt prior AR model and keeping low loss latent variable unaltered. New samples are drawn only when their latent NLL is above a threshold $\lambda_p$. **Fig. 2** illustrates this process (that we refer to as "Latent Space Restoration"). The restoration image is then generated with the decoder network, and the residual image is computed as $|X - \text{Restoration}|$.

Multiple restorations (j ∈ 1,2,...,S) are generated for each test image to reduce variance in the anomaly estimation. The multiple residual images are consolidated using a weighting factor $w_j$ defined as: $w_j = \text{softmax}(k / \sum_i^P |Y^i - X_j^i|)$, where $k$ is a *softmax* temperature parameter and the sum is over all the image pixels $P$. $w$ reduces the weight of restorations which have lost consistency. The final consolidated pixel-wise anomaly score ($AS_{\text{pixel}}$) is estimated as the weighted mean of all residuals:

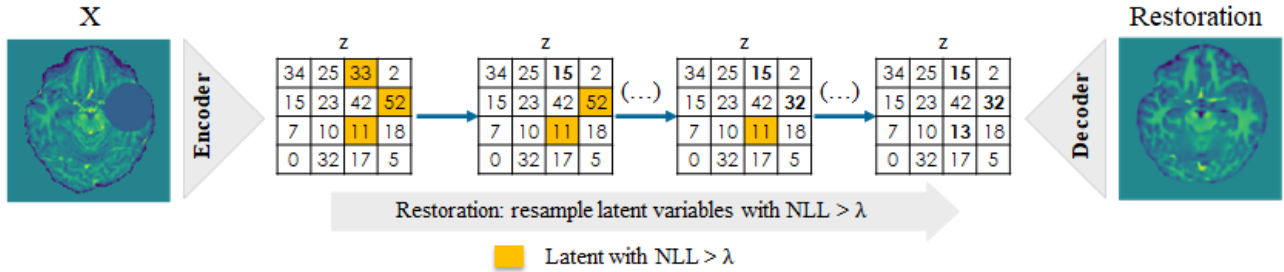

**Fig. 2.** Illustration of Latent Space Restoration process.

$$AS_{pixel} = \sum_{j}^{S} w_j |Y - X_j|$$

Finally, $AS_{pixel}$ scores are smoothed using a 3x3 MinPooling filter followed by a 7x7 AveragePooling filter.

## 3. EXPERIMENTS

### 3.1. Datasets

MOOD challenge [7] datasets have been used to train and evaluate the proposed method. They consist of
- Brain MR: 800 scans obtained from the Human Connectome Project (HCP) dataset [8]. HCP incorporates only young healthy participants;
- Abdominal CT: 550 normal scans from [9].

The challenge test set is kept confidential. However, images from 4 subjects for each dataset with added synthetic anomalies were provided as validation set and used for hyper-parameter tunning. Results listed in the following section correspond to this validation set.

Both datasets are pre-processed according to guidelines from challenge organizers. Images were resized to obtain axials slices of 160x160 pixels. Brain images were normalized to have zero mean and unit standard deviation subject-wise. Image augmentation used in the training set included elastic transforms, gaussian blur, random scale and rotations, random brightness, contrast adjustment and gaussian noise.

### 3.2. Implementation details

Our VQ-VAE includes 5 blocks, each composed of 4 residual blocks and a downsample/upsample operator. The latent space in the brain dataset was set to 20x20 with an embedding space with 128 keys and 256 dimensions. The latent space in the abdominal dataset was set to 10x10. L1 distance was used as reconstruction loss. Additionally, dropout with 0.1 probability is used during training.

PixelSNAIL network consists of 4 blocks, each with 4 residual blocks and a self-attention module. Latent probability distribution was conditioned on the axial slice position. Consequently, in order to estimate a sample probability, the AR model receives as input not only the encoded sample but also the position of the slice within the volume, encoded in a variable in the range [-0.5,0.5]. Dropout with 0.1 probability is used during training.

Network architectures and training procedures were implemented in PyTorch and made openly available in https://github.com/snavalm/lsr_mood_challenge_2020/.

Adam optimizer (with parameters $\beta_1=0.9$, $\beta_2=0.999$, $\varepsilon=10^{-8}$) and learning rate of $10^{-4}$ were used to train both VQ-VAE and PixelSNAIL networks. A batch size of 64 was used in both networks. Batches were created by combining 8 random slices from 8 volumes. VQ-VAE was first trained. The trained encoder was then used to generate the latent variables fed to PixelSNAIL, which was consequently trained. Networks were trained on a single Nvidia GTX1070.

Finally, $\lambda_s$ and $\lambda_p$ thresholds were adjusted using the validation set provided (slice-wise performance was used for $\lambda_s$). We used $\lambda_s = 7$ and $\lambda_p = 5$, corresponding to percentiles 98 and 90 respectively in the validation set. In the pixel-wise score, we found that a lower threshold incentivizes more variance in reconstructions which improved results. $S = 15$ restorations was also heuristically determined.

We compare our method to a standard VAE with the same architecture as the VQ-VAE (5 downsample/upsample blocks, each with 4 residual blocks). A dense layer is incorporated as the final layer of the encoder to define a 128 latent space. VAE loss is used as sample-wise AS and reconstruction as pixel-wise AS.

## 4. RESULTS AND DISCUSSION

Since only 4 volumes are provided for each dataset, we approximated the sample-wise performance using slice-wise performance metrics. Slice and pixel-wise results for our method are summarized in **Table 1**. Area under receiver operating characteristics curve (AUROC) and average precision (AP) are reported for sample and pixel wise scores. In pixel-wise score we additionally evaluated the Dice similarity coefficient (DSC) by identifying abnormal pixels with an AS threshold. We include examples of the sample-wise scores assigned by our method in **Fig. 3**. For pixel-wise scores, **Fig. 4** shows one validation image, 2 of the 15 restorations generated, residuals and final anomaly score.

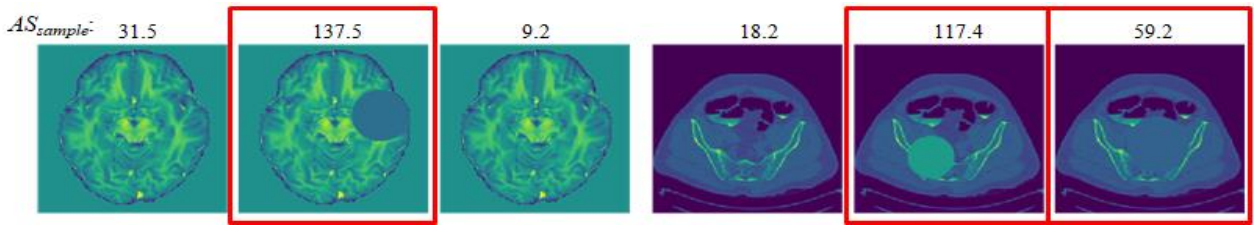

**Fig. 4.** Test images and corresponding sample-wise anomaly scores. Abnormal images are highlighted in red.

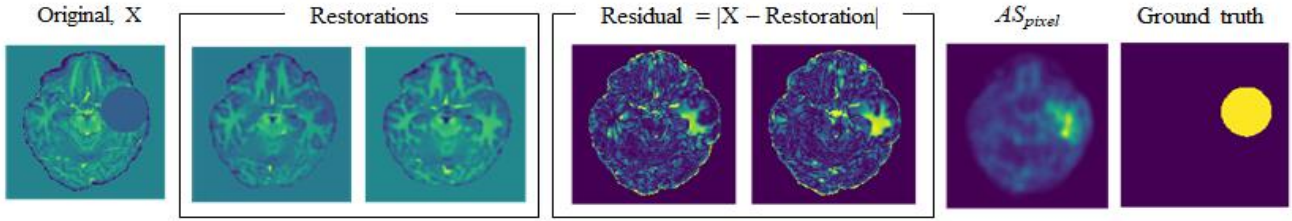

**Fig. 3.** Visualization of restorations and pixel-wise score

|  | $AS_{sample}$ | | $AS_{pixel}$ | | |
|---|---|---|---|---|---|
|  | AUROC | AP | DSC | AUROC | AP |
| *Brain dataset* | | | | | |
| VQ-VAE (ours) | **0.97** | **0.92** | **0.79** | **0.99** | **0.81** |
| VAE | 0.90 | 0.82 | 0.70 | 0.98 | 0.72 |
| *Abdominal dataset* | | | | | |
| VQ-VAE (ours) | **0.83** | **0.73** | **0.57** | **0.98** | **0.57** |
| VAE | 0.65 | 0.48 | 0.29 | 0.93 | 0.23 |

**Table 1.** Comparative of slice and pixel-wise performance.

The obtained results suggest that our approach outperforms a standard VAE method. Pixel-wise results are superior in brain images compared to abdominal (probably due to the higher variance in the abdominal dataset). We also observed that the method is sensitive to the pixel intensity of the anomaly. Anomalies with intensities near the expected intensities are often missed. This can be due to the anomaly scored being calculated as the residual of pixel intensities. Alternative scores will be evaluated in the future.

## 5. CONCLUSIONS

We presented a novel unsupervised anomaly detection and localization method based on VQ-VAEs that improves results upon an existing standard VAE approach. In the MOOD challenge, our approach achieved 2nd and 3rd position in sample and pixel-wise respectively, only surpassed by non-VAE-based methods. In the future, we intend to evaluate our approach in a broader range of datasets and medical anomalies to better assess its robustness and usefulness in a realistic scenario.

## 6. COMPLIANCE WITH ETHICAL STANDARDS

This research study was conducted retrospectively using human subject data made available in open access by [9]. Ethical approval was not required as confirmed by the license attached to the open access data.

## 7. ACKNOWLEDGEMENTS

The authors declare no conflicts of interest.